\documentclass[sigconf]{acmart}
\usepackage{arydshln}
\usepackage{subfigure}
\usepackage{cleveref}
\crefname{section}{§}{§}
\crefname{appendix}{Appendix}{§}

\renewcommand{\vec}[1]{\boldsymbol{#1}} 

\AtBeginDocument{%
  \providecommand\BibTeX{{%
    \normalfont B\kern-0.5em{\scshape i\kern-0.25em b}\kern-0.8em\TeX}}}

\begin{document}

\title{Dialog-to-Actions: Building Task-Oriented Dialogue System via Action-Level Generation}


\author{Yuncheng Hua}
\author{Xiangyu Xi}
\authornote{Yuncheng Hua and Xiangyu Xi contributed equally to this research.}
\orcid{0000-0002-4238-5071}
\affiliation{%
  \country{Meituan Group, China}
}
\email{devin.hua@monash.edu}
\email{xixy10@foxmail.com}


\author{Zheng Jiang}
\affiliation{%
  \country{Meituan Group, China}
}
\email{zjiang@seu.edu.cn}

\author{Guanwei Zhang}
\affiliation{%
  \country{Meituan Group, China}
}
 \email{zhangguanwei@meituan.com}

\author{Chaobo Sun}
\affiliation{%
  \country{Meituan Group, China}
}
 \email{sunchaobo@meituan.com}

\author{Guanglu Wan}
\affiliation{%
  \country{Meituan Group, China}
}
\email{wanguanglu@meituan.com}

\author{Wei Ye}
\affiliation{%
  \country{Peking University, China}
  \country{China}
  }
\email{wye@pku.edu.cn}

\begin{abstract}
  End-to-end generation-based approaches have been investigated and applied in task-oriented dialogue systems. However, in industrial scenarios, existing methods face the bottlenecks of reliability (e.g., domain-inconsistent responses, repetition problem, etc) and efficiency (e.g., long computation time, etc). In this paper, we propose a task-oriented dialogue system via action-level generation. 
  Specifically, we first construct dialogue actions from large-scale dialogues and represent each natural language (NL) response as a sequence of dialogue actions.
  Further, we train a Sequence-to-Sequence model which takes the dialogue history as the input and outputs a sequence of dialogue actions.
  The generated dialogue actions are transformed into verbal responses.
  Experimental results show that our light-weighted method achieves competitive performance, and has the advantage of reliability and efficiency.
\end{abstract}





\maketitle

\section{Introduction}
Recently, the end-to-end generation-based methods that directly output appropriate NL responses or API calls have been deeply investigated in task-oriented chatbots \cite{byrne-etal-2021-tickettalk,yang2021ubar,lin2021bitod,jang2021gpt,thoppilan2022lamda}, and have been proven valuable for real-world business, especially after-sale customer services \cite{li2017alime,yan2017building,zhu2019case,lee-etal-2019-convlab,zhu-etal-2020-convlab,song-etal-2021-emotional,acharya-etal-2021-alexa,sun-etal-2021-adding,taskflow}.
Based on the large-scale pre-trained language models \cite{radford2019language,raffel2020exploring}, generation-based methods have the advantage of simpler architecture and anthropomorphic interaction.
Despite the significant progress, we find these token-level generation methods suffer from the following two limitations in practical scenarios.

\textbf{1. The token-level generation methods have limited reliability, which is essential for industrial task-oriented dialogue systems.}
Due to the pre-trained language models' characteristics, the models may generate responses that are learned from the pre-training corpus. In certain cases, such responses are meaningless and not semantically incoherent with the current business domain, interrupting online interaction.
Worse still, the models occasionally get stuck in generating repetitive responses across multiple turns (e.g., repeatedly enquiring the users for the same information).
Above issues are also widely observed by other researchers \cite{see2019makes,fu2021theoretical} and practitioners.\footnote{\url{https://github.com/microsoft/DialoGPT/issues/45}}

\textbf{2. The token-level generation methods may fail to meet the efficiency requirement of the industrial systems, especially with large decoding steps.}
The long computation time of the token-level generation models leads to unacceptable response latency of online dialogue systems, especially when the model generates a sentence of length that exceeds a threshold (e.g., 1,544 ms of T5 for a sentence of 30 words, as Figure \ref{fig:computation_time} shows).
Owing to the latency problem, a large number of service requests may be suspended or blocked during the peak period.
Also, the computation resources (e.g., GPUs) required by the aforementioned systems might be unaffordable for small companies.


\begin{figure*}[!hbt]
    \centering
    \includegraphics[scale=0.15]{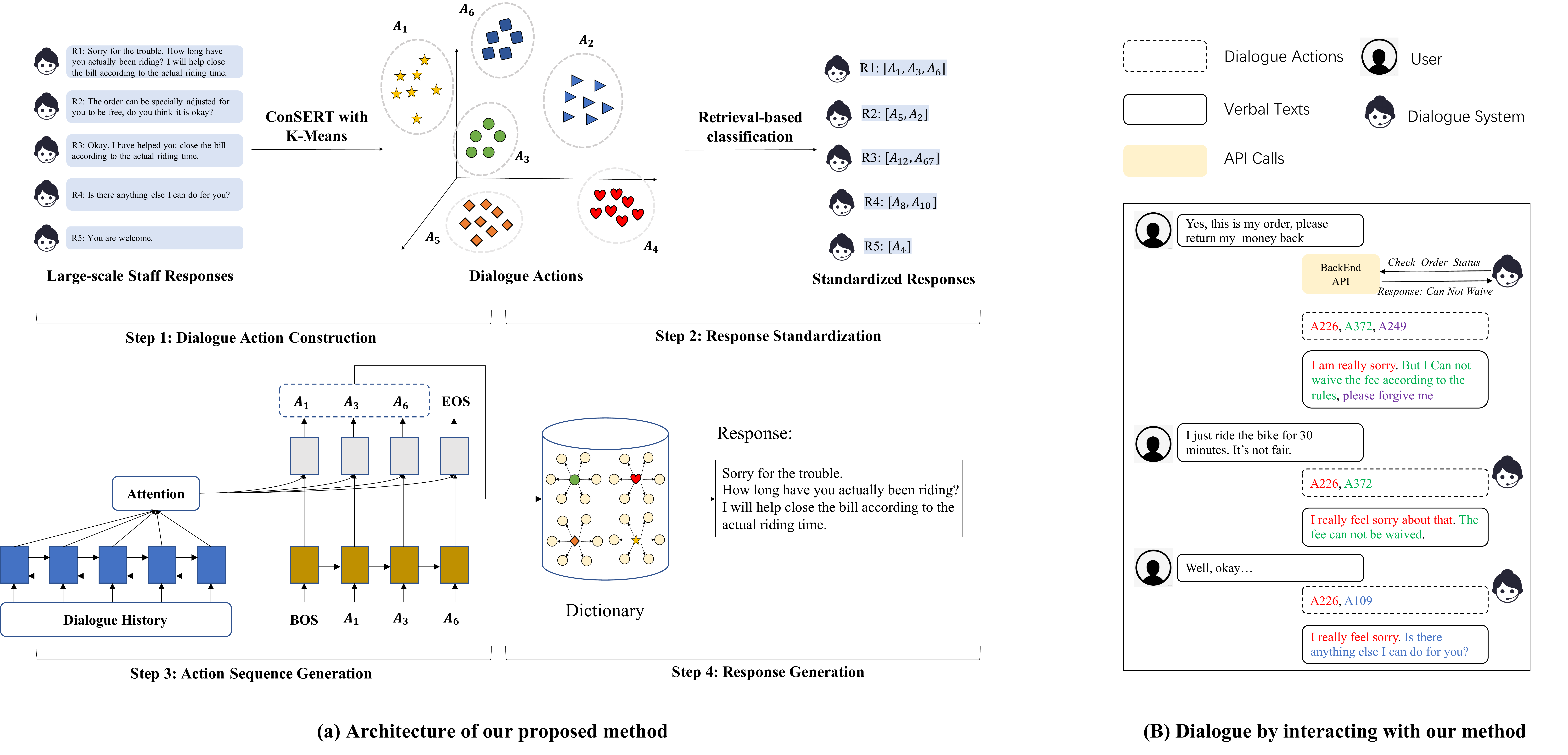}
    \caption{The system architecture and dialogue sample. In (b), the dialogue action and corresponding utterance segment are marked by the same color (e.g., ``\textcolor{red}{A226}'' and ``\textcolor{red}{I am really sorry}'').}
    \label{fig:mdoel}
    \vspace{-4mm}
\end{figure*}

To address the above two problems, in this paper, we propose a task-oriented dialogue system based on the action-level generation method.
Inspired by \citet{taskflow}, we represent responses with \textbf{Dialogue Actions}, i.e., a class of the responses with unique and identical semantic meaning that can be automatically obtained by clustering.
While \citet{taskflow} directly treats a whole response as a specific dialogue action, we split one response into multiple segments \cite{jin2004segmentation} and each segment can be mapped to a dialogue action.
In this way, each response is represented as a sequence of dialogue actions.
Given the dialogue context, a Seq2Seq model with an action-level recurrent decoder is used to generate the sequence of dialogue actions.
Further, a frequency-based sampling method is used to compose the final response, based on the generated sequence of dialogue actions.
Since the core component of our approach is the generation model which takes the \emph{dialogue context} as the inputs and outputs \emph{actions}, our method is named as \underline{D}ialog-\underline{T}o-\underline{A}ctions (abbr. \textbf{DTA}).
Compared with existing token-level generation-based systems, our DTA has the advantage of 
1) reliability, since the generated natural language responses derive from the predefined dialogue actions;
2) efficiency, since the decoding space (i.e., dialogue actions) and the decoding steps are much smaller.

\section{Framework Description}

\subsection{Overview}
We follow the workflow employed in the previous end-to-end task-oriented dialogue systems \cite{byrne-etal-2021-tickettalk,lin2021bitod}, where the system takes the dialogue history as input, and generates a text string that either serves as a verbal staff response to the user or API calls (e.g., information inquiry, action execution, etc).
When an API is invoked, the information returned from the API will be incorporated into the system's next response.
A dialogue sample following such system interaction life cycle can be found in Figure \ref{fig:mdoel} (b).


The key idea of our work is to generate dialogue actions and then compose a verbal response. To do so, we first construct dialogue actions from large-scale dialogues (Step 1) and represent each response as a sequence of dialogue actions (Step 2), as Figure \ref{fig:mdoel} (a) shows.
A Seq2Seq model with an action-level recurrent decoder is utilized to generate dialogue actions (Step 3), and the generated actions are further used to compose the verbal response (Step 4). 
We exemplify using the after-sale customer service of electric bike rental business, where users and staffs communicate online through text messages.
The technical details are introduced as follows.



\subsection{Step 1: Dialogue Action Construction\label{subsec:dialog_action_construct}}

A dialogue action refers to a cluster of utterances or utterance fragments that share identical semantic meaning and represent a common communicative intention, for instance making a request or querying information.
\citet{taskflow} views a group of utterances with identical semantic information as dialogue action and selects a response corresponding to a specific staff action.
However, the oversimplified setting, i.e., abstracting a whole utterance into an action, leads to relatively limited expressiveness and scalability.
%
To make the responses more targeted and flexible, we construct dialogue actions based on utterance segments (of staff) rather than utterances. Specifically, each utterance is divided into multiple segments by a rule-based approach \cite{jin2004segmentation}. Further, following \citet{taskflow}, we exploit a two-stage method to cluster the segments.
Specifically, ConSERT \cite{yan-etal-2021-consert} is utilized to generate representations for each utterance segment, and K-means is then applied to cluster the segments. We choose the number of clusters $K$ empirically to balance the purity and the number of the clusters, and treat each cluster of segments as a dialogue action (e.g., $A_1$ and $A_2$ in Figure \ref{fig:mdoel} (a)).





\subsection{Step 2: Response Standardization\label{subsec:standard}} 
\begin{figure}[hbt]
    \centering
    \includegraphics[scale=0.28]{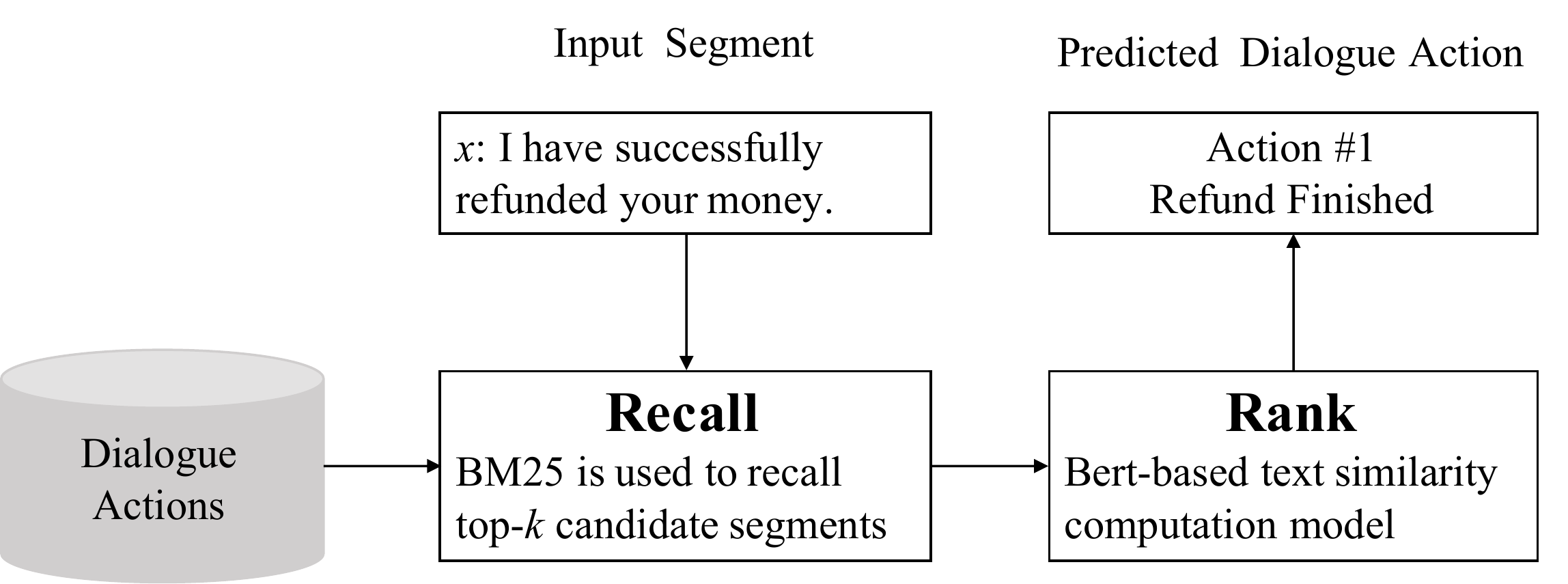}
    \caption{The workflow of response standardization.}
    \label{fig:Standardization}
    \vspace{-4mm}
\end{figure}

Response standardization aims to standardize the responses (from the large-scale dialogues) by mapping each response to a sequence of dialogue actions.
Following \citet{yu-etal-2021-shot}, we exploit a retrieval-based method, which retrieves clustered segments that are most similar to the given input utterance segment and label the input based on the corresponding clusters.
As Figure \ref{fig:Standardization} shows, given an input segment $x$, we use BM25 to recall top $k$ segments $\{u_1,...,u_k\}$ from all clustered segments.
Further, we exploit a BERT-based text similarity computation model $S$ to rerank the $k$ segments and select the segment $\hat u$ with the highest similarity to $x$, denoting as:
\begin{equation}
    \hat u = \underset{u_i \in \{u_1,...,u_k\}}{\mathrm{argmax}} S(x,u_i)
\end{equation}
where $S(x,u_i)$ refers to the similarity between $x$ and $u_i$.
$x$ is then annotated with the dialogue action $A_i$ that $\hat u$ belongs to.

Furthermore, we record the correspondence between the dialogue actions and utterance segments, as well as the frequencies of utterance segments in the dialogues.
Specifically, we employ a key-value dictionary $\mathfrak{D}$, in which a key refers to a dialogue action $A_i$ while its value is a nested dictionary where the mapping relationship between the unique segments $\{x_1,...,x_n\}$ and $A_i$ and the segments' occurrence frequencies $\{f_1,...,f_n\}$ are recorded. 
The dictionary $\mathfrak{D}$ is used for composing the verbal response (Step 4).

\subsection{Step 3: Action Sequence Prediction\label{subsec:action_seq_pred}}

Given a dialogue $\mathcal{D}=\{U_1,S_1,...,U_T,S_T\}$ as a set of utterances exchanged between user ($U_i$) and staff ($S_i$) alternatively, through above carefully-designed steps, each staff utterance $S_i$ is represented as an action sequence $\mathcal{A}_{S_{i}}$.
At the $m$-th turn, given a dialogue history $\mathcal{H}_m=\{U_{m-w},S_{m-w},...,S_{m-1},U_m\}$, we propose a Seq2Seq model to produce staff responses $S_m$. We first resort to the Seq2Seq model to output an action sequence of $\mathcal{A}_{S_{m}}=(A_1^{S_{m}},A_2^{S_{m}},...,A_k^{S_{m}})$, where $k$ denotes the length of the action sequence, and then use the action sequence to form verbal response (in \cref{subsec:response_generate}). \\
\textbf{Encoder}
Given the dialogue history $\mathcal{H}_m$, we sequentially concatenate all the utterances in $\mathcal{H}_m$ and each staff utterance's corresponding action sequence $\mathcal{A}_{S_{i}}$, forming a token sequence $(w_{1},...,w_{n})$.
The Bi-LSTM model is used to encode the token sequence into a sequence of continuous representations $\vec{H}$:
\begin{equation}
    \vec{H} =(\vec{h}_1,...,\vec{h}_n)= {\rm BiLSTM}(w_{1},...,w_{n}) 
\end{equation}
\\
\textbf{Decoder}
Considering the efficiency requirement and small decoding space, we use the Luong attention method and employ LSTM as the decoder to calculate hidden state $\vec{s_{t}}$ at time-step $t$ as follows:
\begin{equation}
    \vec{s_t} = {\rm LSTM}(\vec{s_{t-1}},g(\vec{y_{t-1}}),\vec{c_{t-1}})
\end{equation}
where $\vec{y_{t-1}}$ denotes the probability distribution over dialogue action space at step $t$-1 and $g(\vec{y_{t-1}})$ denotes the action has the highest probability. 
After obtaining hidden state $\vec{s_t}$ and context vector $\vec{c_t}$, we generate probability distribution at time-step $t$ as follows:
\begin{equation}
\vec{y_{t}} = {\rm Softmax} (\vec{W_d}[\vec{s_t};\vec{c_t}])
\end{equation}
where $\vec{W_d}$ is weight parameter. Given the ground-truth label $y_t$ at time-step $t$, we use $p(y_t|y_{<t}, \mathcal{H}_m)$ to denote the cross-entropy loss at step $t$ where $y_{<t}$ denotes the previously-generated actions.

The optimization objective is defined as:
\begin{equation}
    \begin{split}
    L_{Gen} & = -\sum_{\mathcal{D}\in \mathcal{C}}\sum_{\mathcal{H}_m\in \mathcal{D}}\sum_{t=1}^{l_m} p(y_t|y_{<t}, \mathcal{H}_m)
    \end{split}
\end{equation}
where $\mathcal{C}$ denotes the set of dialogues and $l_m$ denotes the length of the dialogue action sequence at the $m$-th turn of dialogue $\mathcal{D}$.


\subsection{Step 4: Response Generation\label{subsec:response_generate}}
Based on the action sequence generated in \cref{subsec:action_seq_pred},
we compose the verbal response by selecting an utterance segment for each action and combining the segments sequentially.
Considering the segments with higher frequencies are more likely to be the formal utterance that staff commonly use, we sample the segments from $\mathfrak{D}$ (built in \cref{subsec:standard}) following the principle that the higher the frequency, the more likely the segment to be selected.
By doing this, we ensure the quality as well as the diversity of the verbal responses.

\section{Experiments}
\subsection{Experimental Settings}
\subsubsection{Dataset}
We perform an experiment with a Chinese online after-sale customer service of electric bike rental business. In this scenario, the users may finish riding but forgot to lock the bike, and thus request the staff to remotely lock the bike and reduce the fees. The staff is required to judge whether the fee can be reduced by checking the status of the order via the back-end APIs.
We collect the user-staff dialogues from the logs of online services for a week. The data statistics are shown in Table \ref{tab:dataset}.
The dialogues are randomly split into train, dev, and test sets with a ratio of 8:1:1.
We construct 1,420 dialogue actions (\cref{subsec:dialog_action_construct}), and each API call is treated as a dialogue action.
The dataset will be released online.

\begin{table}[hbt]
    \centering
    \begin{tabular}{lr}
    \hline
    \textbf{STAT TYPE} & \textbf{VALUE} \\
    \hline
    Dialogs & 8,363\\
    Total turns & 55,576\\
    Avg. turns per dialog & 6.65\\
    Dialogue Actions & 1,420 \\
    \hline
    \end{tabular}
    \caption{Data statistics.}
    \label{tab:dataset}
    \vspace{-6mm}
\end{table}

\subsubsection{Baselines}
\label{baselines}
To evaluate the effectiveness and efficiency of our method, we compare it with the following state-of-the-art baselines:
(1) \textbf{LSTM} which exploits a classical LSTM-based sequence-to-sequence architecture \cite{sutskever2014sequence};
(2) \textbf{Transformer} which uses Transformer \cite{vaswani2017attention} as encoder and decoder;
(3) \textbf{CDiaGPT} which is a GPT model pre-trained on a large-scale Chinese conversation dataset \cite{wang2020chinese};
(4) \textbf{T5} which is a Text-to-Text Transfer Transformer (T5) model \cite{raffel2020exploring} pretrained with the CLUE Corpu.


\subsubsection{Evaluation Metrics}

To comprehensively evaluate the effectiveness of different models, we perform both offline evaluation (i.e., on the dataset) and online evaluation (i.e., online A/B testing).

\textbf{Offline Evaluation} The models take a specific ground-truth conversation history (i.e., context) as input and generate a response. Following \citet{byrne-etal-2021-tickettalk}, we report the BLEU-4 score of each model in the test set. Considering the API calls are highly important, we observe the generated API calls in each turn and report the macro Precision (P), Recall (R), and F1-Score (F1) of API calls.

\textbf{Online Evaluation} Following \citet{taskflow}, we deploy the models online and perform A/B testing. For each model, 120 dialogues are randomly sampled. The annotators who possess domain knowledge are required to perform the satisfaction assessment by grading each dialogue as ``Low'', ``Medium'', or ``High'' satisfaction degree. \footnote{The grading criteria can be summarized as follows: (i) Score ``Low'' denotes that Chatbot can not handle user requirements correctly. (ii) Score ``Middle'' denotes that Chatbot can handle user requirements correctly, but may generate a disfluent or incomplete response. (iii) Score ``High'' denotes that Chatbot can handle user requirements correctly and complete the conversation perfectly.}

\subsection{Main Results}
The offline evaluation and online evaluation are shown in Table \ref{tab:api_prediction} and \ref{tab:human_evaluation} respectively, from which we have the following observations:
(1) Large-scale pre-trained language models significantly improve the performance of token-level generation models. For example, compared with the plain LSTM model, T5 achieves an absolute improvement of 26.20\% for the BLEU-4 score and 4.98\% for F1.
(2) Compared with the CDiaGPT and the T5 models, our light-weighted DTA achieves competitive performance in the offline evaluation, and earns the highest satisfaction rating in the online evaluation, verifying the effectiveness of our proposed method.


\begin{table}[hbt]
    \centering
    \begin{tabular}{lccc}
    \toprule
    Model &  Low & Medium & High\\
    \hline
    LSTM & 30.00 & 29.17 & 40.83\\
    Transformer & 27.50 & 28.33 & 44.17\\
    CDiaGPT & 12.50  & 13.33 & 74.17 \\
    T5 & 5.83 & 15.83 & 78.33 \\
    DTA & 8.33  & 12.50 & 79.17 \\
    \bottomrule
    \end{tabular}
    \caption{Statistical Results of Online Evaluation (\%)}
    \label{tab:human_evaluation}
    \vspace{-6mm}
\end{table}

\begin{table}[hbt]
  \centering
  \begin{tabular}{lcccc}
      \toprule
        \textbf{Model}& \textbf{BLEU-4} & \textbf{P} & \textbf{R} & \textbf{F1} \\
        \hline
        LSTM & 20.62&66.16 &78.84 & 71.72\\
        Transformer & 26.21& 62.59&75.90&64.74 \\
        CDiaGPT & 42.54 & \textbf{71.18} & 86.43 & 77.11\\
        T5 & \textbf{46.83}  & 70.91 & 87.25 & 76.70 \\
        DTA & 44.82 & 68.03 & \textbf{90.80} & \textbf{77.74}\\
      \bottomrule
  \end{tabular}
  \caption{Statistical Results of Offline Evaluation (\%).}
  \label{tab:api_prediction}
  \vspace{-6mm}
\end{table}
\subsection{In-Depth Analysis}
\subsubsection{Effect of Efficiency Issue}
\begin{figure}[!hbt]
    \centering
    \includegraphics[scale=0.12]{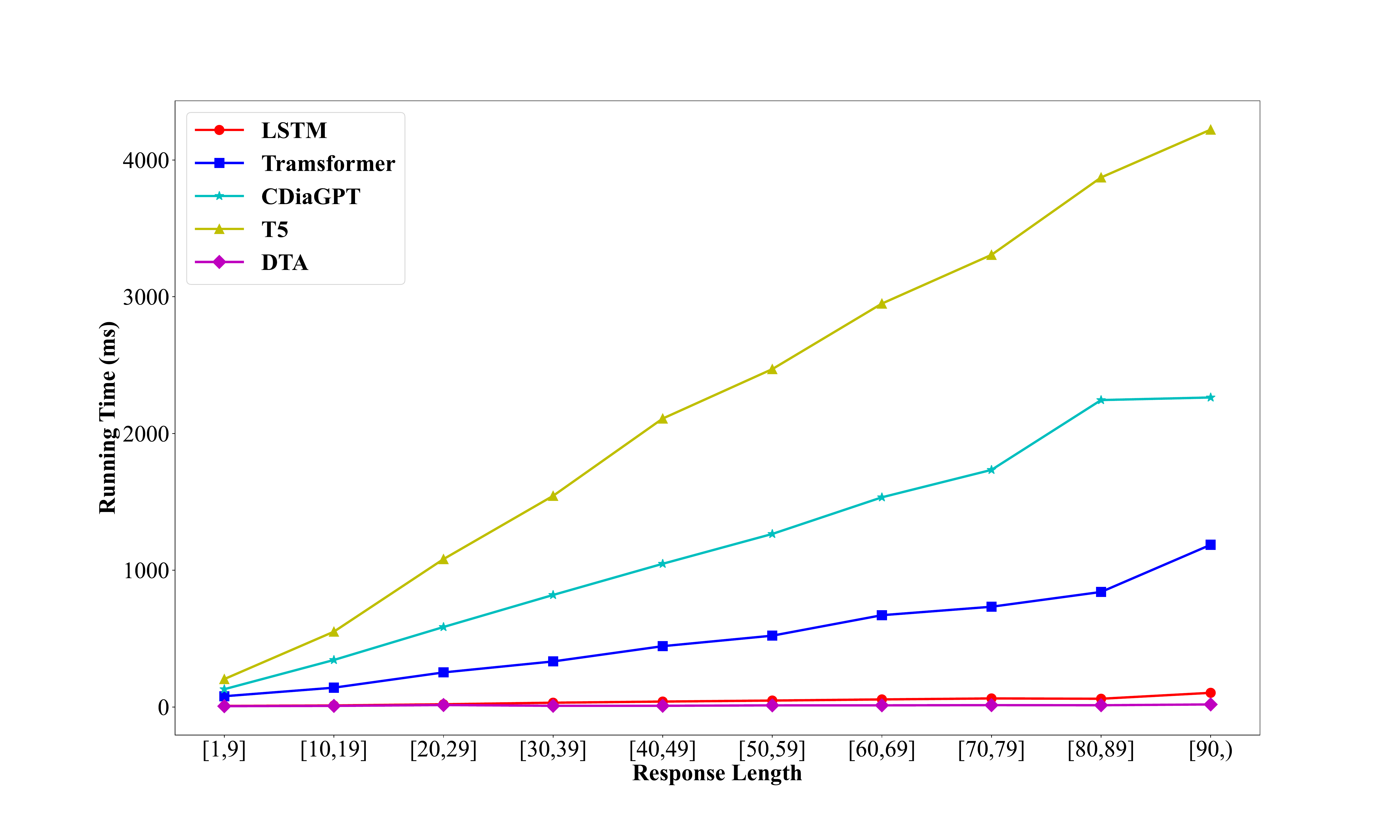}
    \caption{The computation time of different models.}
    \label{fig:computation_time}
\end{figure}

To investigate the effect of efficiency issue, we collect each model's computation time for processing the test samples under the same infrastructure (i.e., Tesla v100, 32GB RAM size, etc). Considering the computation time is highly correlated with the decoding steps, we first divide the generated responses into 10 subsets based on the response length, and then calculate the average computation time of each subset, as Figure \ref{fig:computation_time} shows.
We can observe that:
(1) Despite the comparable performance, DTA has a significant advantage in computation efficiency over other models (e.g., 3.37 ms of DTA v.s. 1265.56 ms of CDiaGPT v.s. 2470.69 ms of T5 for subset ``[50, 59]'').
(2) DTA outperforms other models more significantly with longer responses. The reason is that the decoding steps of the token-level generation models are identical to the response length, while DTA performs action-level decoding.
The above observation verifies that our system provides an effective solution to build online dialogue services with limited computation resources.

\subsubsection{Effect of Reliability Issue}

We investigate the effect of reliability issue by quantitatively inspecting the repetition problem.
Specifically, we calculate the Jaccard index of responses of each turn and the previous turns \cite{costa2021further}.
The average Jaccard index of each model is shown in Table \ref{tab:repetition}, from which we can observe that:
(1) The Jaccard index of human response is the smallest, indicating that existing models have room for improvement in terms of the repetition problem.
(2) Our method has a much smaller Jaccard index than CDiaGPT and T5. The action sequence generation, together with the sampling strategy, can effectively alleviate the repetition problem.

\begin{table}[!hbt]
    \centering
    \begin{tabular}{lc}
    \toprule
    Model   &  Jaccard Index\\
    \hline
    Human Response & 0.129 \\
    \hline
    CDiaGPT &  0.214 \\
    T5 & 0.207\\
    DTA & 0.142 \\
    \bottomrule
    \end{tabular}
    \caption{Jaccard Index of different models.}
    \label{tab:repetition}
    \vspace{-6mm}
\end{table}

A concrete online dialogue is shown in Figure \ref{fig:mdoel} (b), where the CDiaGPT generates exactly the same responses.
Though DTA generates similar action sequences (e.g., combinations of Actions \textcolor{red}{A226}, \textcolor{green}{A372}, \textcolor{purple}{A249} and \textcolor{blue}{A109}), there are much fewer cases where the action sequence exactly repeats previous turns. The small differences in action sequences can lead to large changes in verbal responses.
Besides, the sampling mechanism ensures that the same action in different turns corresponds to different segments (e.g., \textcolor{red}{A226} in three turns), which further enables DTA with better diversity.

\section{Conclusion}

In this paper, we propose a task-oriented dialogue system via action-level generation. An effective framework is proposed to build the generation model from the large-scale dialogues with minimum manual effort.
The experimental analyses demonstrate our system's capability of tackling the reliability and the efficiency problems encountered with the existing end-to-end generation methods.
In the future, we are interested in exploring an integrated system that unifies the discrete modules in DTA in an end-to-end architecture.


\section{Presenter BIOGRAPHY}
Presenter: Yuncheng Hua. He is an algorithm engineer at Meituan, focusing on researching and building dialogue systems.

\section{Company Portrait}
Meituan is China's leading shopping platform for locally found consumer products and retail services including entertainment, dining, delivery, travel and other services.



\bibliographystyle{ACM-Reference-Format}
\bibliography{sample-base}

\end{document}